\documentclass[a4paper]{article}

\usepackage{INTERSPEECH2022}

\usepackage{times}
\usepackage[utf8]{inputenc}
\usepackage[T1]{fontenc}
\usepackage{graphicx}
\usepackage{hyperref}
\hypersetup{colorlinks=true, allcolors=blue}
\usepackage{comment}
\usepackage[subtle]{savetrees}
\usepackage[ss]{cmll}
\usepackage{xcolor}
\usepackage{soul}

\usepackage{siunitx}

\usepackage[english]{babel} 

\title{Toward Low-Cost End-to-End Spoken Language Understanding}

\name{Marco Dinarelli, Marco Naguib, François Portet}

\address{
  Univ. Grenoble Alpes, CNRS, Grenoble INP, LIG, 38000 Grenoble, France}
\email{(marco.dinarelli|francois.portet)@univ-grenoble-alpes.fr, marco.naguib@hotmail.com}

\begin{document}

\maketitle

\begin{abstract}
Recent advances in spoken language understanding benefited from Self-Supervised models trained on large speech corpora. For French, the LeBenchmark project has made such models available and has led to impressive progress on several tasks including spoken language understanding. These advances have a non-negligible cost in terms of computation time and energy consumption. 
In this paper, we compare several learning strategies trying to reduce such cost while keeping competitive performance. At the same time we propose an extensive analysis where we measure the cost of our models in terms of training time and electric energy consumption, hopefully promoting a comprehensive evaluation procedure. The experiments are performed on the FSC and MEDIA corpora, and show that it is possible to reduce the learning cost while maintaining state-of-the-art performance and using SSL models.
\end{abstract}

\section{Introduction}

Spoken Language Understanding (SLU) aims at extracting semantic representations from speech signals containing sentences in natural language \cite{DeMori1997:SDBook}.
Classical approaches to SLU used a cascade model made of an Automatic Speech Recognizer (ASR) feeding a Natural Language Understanding (NLU) module \cite{Raymond2006:FST-SEM,Hahn.etAL-SLUJournal-2010,caubriere:hal-02465899,ghannay:hal-03372494}.
Neural networks led to large advances for \emph{end-to-end} SLU systems \cite{DBLP:journals/corr/abs-1802-08395,desot:hal-02464393,lugosch2019speech,DBLP:journals/corr/abs-1906-07601,dinarelli2020data,pelloin:hal-03128163}, 
which are preferred to cascade systems, in particular for their ability to reduce error propagation effects and to exploit acoustic components to deduct semantic information \cite{DESOT2022101369}. \\
Though some approaches proposed full end-to-end learning of SLU models, e.g. \cite{8268987,desot:hal-02464393,palogiannidi2020endtoend}, most proposals use incremental learning on more and more specific tasks (e.g. ASR) before SLU.
Based on the assumption that an SLU model must learn the lexical representation supporting meaning, \cite{lugosch2019speech} and \cite{dinarelli2020data} propose 
a model trained progressively to transcribe the signal and then to extract the semantics.
In this trend there are also \cite{serdyuk2018endtoend} and \cite{radfar2020endtoend}, which learn first a domain classifier, and then optimize the model for predicting user intent and semantic slots.
Recently end-to-end SLU approaches integrated in the system Self-Supervised Learning (SSL) models trained on huge amount of speech data, e.g. \emph{wav2vec} and \emph{HuBERT} \cite{DBLP:journals/corr/abs-1904-05862,NEURIPS2020_92d1e1eb,DBLP:journals/corr/abs-2106-07447}.
\cite{Lai2020} for example, uses a pre-trained wav2vec model as speech encoder, while the \emph{SUPERB} benchmark \cite{superb} proposes slot-filling and intent classification tasks to evaluate the SSL pre-trained models.
SSL models are now available also for the French community \cite{evain:hal-03317730,evain:hal-03407172}, and led to impressive improvements on downstream tasks, including SLU. \\
While such advances are a great scientific achievement, they come at non-negligible costs in terms of resource consumption \cite{strubell-etal-2019-energy,10.1145/3442188.3445922,parcollet:hal-03190119}.
For instance the largest model in \cite{evain:hal-03407172} is trained during two weeks on $64$ GPUs, and this is even not that expensive compared to larger models or models trained on far more data \cite{conneau2020unsupervised}. We can argue this is still a limited cost as such models are trained once and used for many applications.
However, SSL models are often a component of the whole system, and they need to be fine-tuned on the downstream task to achieve good results.
This practice multiplies costly training phases, which may lead to huge resource consumption. \\
In this paper we aim at analysing and reducing the cost for obtaining competitive performance on SLU tasks while exploiting SSL pre-trained models such as those deployed by \cite{evain:hal-03317730,evain:hal-03407172}.
We propose an analysis for finding a better compromise between performance and energy consumption.
We show the computational cost of our models in terms of different measures, including training time, energy consumption and carbon foorprint.
Our analysis is based on different, yet simple training strategies with respect to those used in \cite{evain:hal-03317730,evain:hal-03407172}.
In addition, for the French task MEDIA \cite{bonneau-maynard-etal-2006-results}, we couple our training strategies with: \emph{i)} a transfer learning strategy using models for a different task \cite{Lefevre2012}; \emph{ii)} an offline fine-tuning phase of a SSL model directly on the SLU target task, instead of the ASR task as proposed in \cite{evain:hal-03407172}.
Such fine-tuning is also performed for the FSC task \cite{lugosch2019speech} using the \emph{XLSR53} SSL model \cite{conneau2020unsupervised}.
While our fine-tuning is relatively expensive with respect to the downstream SLU model learning, it is still cheaper than recent approaches, while allowing for comparable performance.

\begin{table*}[t]
\begin{center}
\scalebox{0.68}{
    \begin{tabular}{l|c|c|c|c|c|c|c}

        \hline
        \multicolumn{8}{c}{Corpus: PortMEDIA, Metric: Concept Error Rate (CER $\shpos$)} \\
        \hline
        \textbf{Train stg} & \textbf{\# params (M)} & \textbf{Input} & \textbf{kWh} (gCO2) & \textbf{kWh/p} & \textbf{Train T} & \textbf{DEV} & \textbf{TEST} \\
        \hline
        \hline
        \multicolumn{8}{|c|}{\textbf{Base features}} \\
        \hline
        3 steps   & 9.83 &   spectro & 4,473 (228) & 0,099 & 36h14' &   35.91  &   \textbf{40.57} \\ 
        2 steps   & 9.83 &   spectro & 2,989 (152) & $\infty$ & 24h14' &   65.80  &   87.32 \\ 
        1 step   & 9.83 &   spectro & 1,708 (87) & $\mathcal{M}_2$ & 15h52' &   59.22  &   68.50 \\
        \hline
        \hline
        3 steps   & 11.76 &   w2v2-fr & 3,983 (203) & 2,235 & 36h22' &   22.17 &   22.51 \\ 
        2 steps   & 11.76 &   w2v2-fr & 2,707 (138) & 1,939 & 24h27' &   21.86  &   23.02 \\ 
        1 step   & 11.76 &   w2v2-fr & 1,815 (93) & $\mathcal{M}_2$ & 18h08' &   25.53  &   23.48 \\
        \hline
        \hline
        \multicolumn{8}{|c|}{\textbf{Fine-tuned features} (+100h x4 GPU $\Rightarrow$ 92.720 kWh / 4729 gCO$_2$)} \\
        \hline
        1 step +1   & 11.76 (+318) &   w2v2-fr slu & 1,214 (62) & - & 11h34' &   21.50  &   22.13 \\ 
        \hline
        \multicolumn{8}{|c|}{\textbf{Literature}} \\
        \hline
        \cite{caubriere:hal-02304597}   & - &   MFCC & - & - & - &   -  &   \textbf{42.3} \\
        \hline
        
    \end{tabular}
}
\end{center}
\caption{Results on the French corpus PortMEDIA. See the text for details.}
\label{tab:PortMEDIA}
\end{table*}

\begin{table*}[t]
\begin{center}
\scalebox{0.68}{
    \begin{tabular}{l|c|c|c|c|c|c|c}

        \hline
        \multicolumn{8}{c}{Corpus: FSC, Metric: Accuracy $\shneg$} \\
        \hline
        \textbf{Train stg} & \textbf{\# params (M)} & \textbf{Input} & \textbf{kWh} (CO2) & \textbf{kWh/p} & \textbf{Train T} & \textbf{DEV} & \textbf{TEST} \\
        \hline
        \hline
        \multicolumn{8}{|c|}{\textbf{Base features}} \\
        \hline
        3 steps   & 7.93 &   spectro & 1.474 (75) & 0.259 & 12h32' &   82.14  &   93.78 \\ 
        2 steps   & 7.93 &   spectro & 0.707 (36) & 0.208 & 5h57' &   78.19  &   91.01 \\ 
        1 step   & 7.93 &   spectro & 0.503 (26) & $\mathcal{M}_2$ & 5h02' &   79.79  &   90.03 \\ 
        \hline
        \hline
        3 steps   & 9.70 &   XLSR-53 & 1.068 (54) & 0.519 & 9h57' &   97.63  &   99.66 \\ 
        2 steps   & 9.70 &   XLSR-53 & 0.950 (48) & 0.564 & 8h34' &   96.57  &   99.39 \\ 
        1 step   & 9.70 &   XLSR-53 & 0.668 (34) & $\mathcal{M}_2$ & 8h08' &   96.66  &   98.89 \\ 
        \hline
        \hline
        \multicolumn{8}{|c|}{\textbf{Fine-tuned features} (+100h x4 GPU $\Rightarrow$ 92.720 kWh / 4729 gCO$_2$)} \\
        \hline
        1 step +1   & 9.70 (+315) &   XLSR-53 slu & 0.365 (19) & - & 3h23' &   99.20  &   \textbf{99.71} \\
        \hline
        \multicolumn{8}{|c|}{\textbf{State of the art}} \\
        \hline
        \cite{lugosch2019speech}   & - &   raw audio & - & - & - &   -  &   98.70 \\
        \cite{DBLP:journals/corr/abs-2102-06283}   & 287$^{(*)}$ &   speech+text & - & - & - &   -  &   \textbf{99.71} \\
        \hline
    \end{tabular}
}
\end{center}
\caption{Results on the English FSC corpus. See the text for details. $^{(*)}$: our own estimation.}
\label{tab:FSC}
\end{table*}

\section{SLU and SSL Models}

In this paper we exploit pre-trained SSL models and we seek to get the best outcome as possible on SLU tasks using the least amount of resources.
We use the SSL \emph{w2v2-fr-7k} model from \textit{LeBenchmark} for French \cite{evain:hal-03407172}, 
which led to the best performance on SLU. We use the multi-language \emph{XLSR-53} model for English \cite{conneau2020unsupervised}.\footnote{We were not able to use the English monolingual SSL model in our Fairseq architecture.} \\
\textbf{SLU models} used in this paper are the same as in \cite{evain:hal-03407172}.\footnote{We downloaded models and systems from \url{https://huggingface.co/LeBenchmark} and \url{https://github.com/LeBenchmark/NeurIPS2021} respectively.}
They are \emph{sequence-to-sequence} models based on LSTMs and attention mechanisms \cite{Hochreiter-1997-LSTM,DBLP:journals/corr/BahdanauCB14}.
The encoder has a similar pyramidal structure as the one proposed in \cite{DBLP:journals/corr/ChanJLV15}, the decoder uses two attention mechanisms, one for attending the encoder's hidden states, and one for attending the decoder's previous predictions, like the self-attention module of Transformers \cite{NIPS2017_3f5ee243}.
All models are trained minimizing the CTC loss \cite{Graves:2006:CTC:1143844.1143891}.
In this work we use SSL models as feature extractors. Features are given as input to SLU models as an alternative to traditional features (e.g. MFCC). This usage enables a cheaper training phase than including the SSL model as a component of the SLU system.\footnote{\scriptsize{Comparing at the same effort, that is: offline feature extraction without fine-tuning vs. SSL model as freezed component of the system; or offline fine-tuning and feature extraction vs. SSL model as fine-tuned component of the system.}}

Models described in \cite{evain:hal-03407172} are learned with three training steps. Each training step uses the model learned at the previous step for initializing the current model's parameters. This strategy is named \emph{3 steps} in this paper, and it consists in: {\bf 1)} training the encoder for ASR; {\bf 2)} training the encoder for SLU; {\bf 3)} training the whole model, encoder and decoder, for SLU.
While this strategy is the most effective, it requires a high training cost. Additionally the best models proposed in \cite{evain:hal-03407172} require an additional training step for supervised fine-tuning of the SSL model on the downstream task. \\
Results of SLU models using input features generated with SSL models are very high, we argue that a comparable performance can be reached with a lower training cost, reducing resource consumption.
With our analyses we would like to reach a better compromise between training cost and model's final performance.
In order to validate our hypothesis we propose two alternative and simple training strategies (indicated as \textbf{Train stg} in our tables): in the \emph{``2 steps''} strategy we perform only steps 2 and 3 of the \emph{3 steps} strategy; in \emph{``1 step''} we train directly the final SLU model full end-to-end.


\section{Evaluation}

\subsection{Data}

We will use mainly the MEDIA corpus for French \cite{bonneau-maynard-etal-2006-results}, which we extensively used in the past \cite{dinarelli09:emnlp,Quarteroni.etAl:Interspeech09,dinarelli09:Interspeech,Dinarelli2010.PhDThesis,DinarelliTellier:CICling:2016,Dupont.etAl:LDRNN:CICling2017,dinarelli:hal-01553830}, used also in \cite{evain:hal-03407172} and allowing thus for direct comparison, and the FSC corpus for English \cite{lugosch2019speech}. \\
The MEDIA corpus focuses on the domain of hotel information and reservation in France.
It is made of $1,250$ human-machine dialogues transcribed and annotated with $76$ semantic concepts.
The corpus is split into $12,908$ utterances ($41.5$h) for training, $1,259$ for development ($3.5$h), and $3,005$ for test ($11.3$h).
For French, in this paper we consider the MEDIA task as the target task, that is the task on which we would like to reach the best possible performance at the lowest possible cost, starting possibly from existing resources such as pre-trained SSL models, but also pre-trained SLU models for other (possibly similar) tasks. \\
In order to reach these experimental conditions and inspired by transfer learning techniques of \cite{caubriere:hal-02304597}, we use also the PORTMEDIA (PM) corpus \cite{Lefevre2012}, focusing on ticket reservation for the \emph{2010 Avignon Festival}.
This corpus has been collected and annotated following the same paradigm as for MEDIA.
PM is also divided in training, development and test splits, composed respectively of $5,900$, $1,400$ and $2,800$ utterances.
This corpus has been annotated with 36 semantic concepts close to the MEDIA concept set: PORTMEDIA and MEDIA share indeed 26 concepts.

In order to have a wider view on SLU results and their cost, we also used the \emph{Fluent Speech Commands} (FSC) corpus \cite{lugosch2019speech}, composed of $30,043$ utterances from $97$ speakers.
Each utterance contains a vocal command for a domotic environment (e.g. \emph{increase heat in the kitchen}), and is annotated with 3 attributes: action, object and localisation.
In the original paper, authors define user intent as the combination of these 3 attributes, and SLU results are computed as accuracy on this combination.
The corpus is split in $23,132$ utterances for training (14.7h), $3,118$ for development (1.9h), and $3,793$ for test (2.4h).
Since results on FSC are particularly high since its release \cite{lugosch2019speech}, we did not perform additional experiments with transfer learning on this task.

For more details on data sets and experimental settings we refer the reader to previous work \cite{evain:hal-03407172,lugosch2019speech}.

\subsection{Evaluating the computational cost}

We evaluate the computational cost of our models measuring: the training time, the electric energy consumption in \emph{kWh} and its conversion into grams of CO$_{2}$ (indicated as gCO$_2$ in the tables). These 2 values are obtained using the \emph{codecarbon} tool\footnote{\url{https://codecarbon.io}}.
Since this tool overestimates the conversion between \emph{kWh} and gCO$_2$, like in \cite{parcollet:hal-03190119} we use the official coefficient of $51$ grams/kWh.\footnote{Available on the \href{https://www.eea.europa.eu/data-and-maps/indicators/overview-of-the-electricity-production-3/assessment}{European Environment Agency website}.}
We also show the cost in \emph{kWh} for gaining one point of the evaluation metric, either Concept Error Rate (CER) or Accuracy (Acc.).
This metric is indicated as \textbf{kWh/p} in the tables.
For computing this value we assume a cheaper model $\mathcal{M}_2$, in terms of resource consumption, obtains worse results than a compared model $\mathcal{M}_1$. Defining then $\text{kWh}(\mathcal{M}_i)$ and $\text{CER}(\mathcal{M}_i)$ respectively the energy consumption and the CER of the model $\mathcal{M}_i$, the value \emph{kWh/p} is defined as $\frac{\text{kWh}(\mathcal{M}_1) - \text{kWh}(\mathcal{M}_2)}{\text{CER}(\mathcal{M}_2) - \text{CER}(\mathcal{M}_1)}$, with the constraint $\text{kWh}(\mathcal{M}_1) \ge \text{kWh}(\mathcal{M}_2)$.
Since $\mathcal{M}_2$ is cheaper, the numerator is positive. Assuming $\mathcal{M}_2$ makes a greater amount of errors than $\mathcal{M}_1$, the denominator is also positive.
Defining analogously $\text{Acc}(\mathcal{M}_i)$ the accuracy of model $\mathcal{M}_i$, we can define \emph{kWh/p} for tasks evaluated through accuracy, like FSC in this work: $\frac{\text{kWh}(\mathcal{M}_1) - \text{kWh}(\mathcal{M}_2)}{\text{Acc}(\mathcal{M}_1) - \text{Acc}(\mathcal{M}_2)}$, again we assume a cheaper model is less effective, and we swapped $\mathcal{M}_1$ and $\mathcal{M}_2$ at the denominator since the higher the accuracy the better.
In the following tables, the value \emph{kWh/p} is given with respect to the cheapest model using the same input features (\textbf{Input} in the tables). Such a model is indicated with $\mathcal{M}_2$ in the same column of the table.
When a model $\mathcal{M}_1$ is worse both in terms of CER (or accuracy) and kWh than the reference model $\mathcal{M}_2$, we use by convention the value $\infty$ for \emph{kWh/p}, meaning that a more expensive training would not lead to any performance improvement.
The interpretation of \emph{kWh/p} should be understood keeping in mind the assumption that a cheaper model is less effective. Then, with given input features, \emph{kWh/p} measures the additional cost in kWh to increase the performance by 1 point. Such cost could be due to using more data and/or a bigger model, leading to a longer training time.
The \emph{kWh/p} metric can be interpreted also, except for special cases where its value would be negative or not defined, the other way around: the electric energy saved if we accept a one point performance drop.

\subsection{Results}

Quantitative results for SLU tasks are evaluated with the Concept Error Rate (CER)\footnote{Computed aligning with the edit distance the gold and the predicted concept sequences.} for PortMEDIA and MEDIA tasks, and with accuracy for FSC.
We add in tables the number of parameters, and the metrics for evaluating the computational cost of our models, introduced in the previous section.

\subsubsection{Preliminary experiments for French}

For reaching experimental conditions where we have resources available in advance for our French target task (MEDIA), we trained SLU models for the PortMEDIA task.
These models will be then used for pre-initializing models for MEDIA.
Results on PortMEDIA are reported in table~\ref{tab:PortMEDIA}.
We trained models with two basic features: spectrograms (indicated with \textbf{spectro}) and with features generated with the SSL model \emph{w2v2-fr 7k} from \emph{LeBenchmark}.
As we can see, the best results are obtained always with the most expensive strategy \emph{3 steps}. However, using as input \emph{w2v2-fr} features, the difference between \emph{3 steps} and \emph{1 step} models is less than 1 CER point, while the latter strategy is quite cheaper both in terms of training time and energy consumption: 54.43\% kWh cost reduction with only 4.31\% performance loss.
Even better CER results can be obtained, and at a lower final cost (69.52\% kWh cost reduction with 1.69\% performance gain), using features computed with the \emph{w2v2-fr} SSL model fine-tuned on the SLU MEDIA task.
Supervised fine-tuning is performed like in \cite{evain:hal-03407172}, the SSL model training is resumed using MEDIA data for training and minimizing the CTC loss with respect to the SLU output (see the appendix~A.2.3 in \cite{evain:hal-03407172}).
These results are shown in the last line of table~\ref{tab:PortMEDIA} (\emph{w2v2-fr slu} features).
These results are marked with \emph{1 step +1} in the column \textbf{Train stg} to keep into account the fine-tuning of the SSL model, which requires 100 hours of training on 4 GPUs (\emph{+100h x4 GPU} in the tables) of a 318M parameters model.
Since this model is never a component of the final SLU model in our experiments, it is indicated in the ``\# params'' column in parenthesis. Based on estimations in \cite{evain:hal-03407172}, we can estimate the cost of our fine-tuning to 92.720 kWh, or 4729 gCO$_2$, these values are reported in table headers.
The fine-tuning cost dominates the cost for training the SLU model.
As we mentioned above however, this fine-tuning is performed only once, on the MEDIA task, and with the purpose of sharing such models as additional resources, like models of \emph{LeBenchmark}.
We made this choice both to save resources overall, and because PortMEDIA SLU models are used to pre-initialize SLU models trained for MEDIA as a transfer learning strategy.
As we anticipated in previous sections thus, using available SSL models it is possible to obtain competitive performance on the SLU task at a lower cost (\emph{3 steps} vs. \emph{1 step} strategies).
Logically, results are even better if a fine-tuned SSL model is available. \\
Thanks to this first set of experiments, we have SLU models available, in addition to SSL models, to train SLU models at a lower cost on our French target task, MEDIA.
In a real application scenario, it would be desirable that such resources, including fine-tuned models, were available in advance, and they are exploited by using the same SLU system for training target SLU models, which is what we do in this work.

\subsubsection{Results on English FSC}

Results on the FSC corpus are given in table~\ref{tab:FSC}. As we can see they follow the same trend as results on PortMEDIA. All models obtain very good results on this task,
 even if we did note perform any fine tuning of parameters on this task, the system is used as in \cite{evain:hal-03407172}, which is tuned on MEDIA.
Beyond that, using features from the \emph{XLSR53} model, we obtain very competitive results even with the model trained full end-to-end (\emph{1 step}), with a 37.45\% kWh cost reduction with respect to the \emph{3 steps} strategy and just 0.77\% performance loss.
Results are even better, and obtained at a lower final cost (3h23' for training, 65.82\% kWh cost reduction and 0.05\% performance gain), using a model trained with features from a SSL model tuned on the FSC SLU output.
Again, fine-tuning time and cost dominate SLU model cost, but it is intended to be done once and for all, and for producing resources that will be made available for avoiding to repeat such process.
Additionally, while we fine-tune a model of 315M parameters on FSC data only, that is 14.7 hours of speech, a state-of-the-art model such as \cite{DBLP:journals/corr/abs-2102-06283} pre-trains ASR and BERT-base models, roughly 287M parameters, on 75k hours of speech, then use such models as components in the final SLU system, which is also fine-tuned on the FSC data.

\begin{table*}[t]
\begin{center}
\scalebox{0.68}{
    \begin{tabular}{l|c|c|c|c|c|c|c}

        \hline
        \multicolumn{8}{c}{Corpus: MEDIA, Metric: Concept Error Rate (CER $\shpos$)} \\
        \hline
        \textbf{Train stg} & \textbf{\# params (M)} & \textbf{Input} & \textbf{kWh} (gCO$_2$) & \textbf{kWh/p} & \textbf{Train T} & \textbf{DEV} & \textbf{TEST} \\
        \hline
        \hline
        \multicolumn{8}{|c|}{\textbf{Base Features}} \\
        \hline
        \cite{evain:hal-03407172} 3 steps & 10.15 & spectro & - & - & \textit{57h} & \textbf{29.07} & \textbf{31.10} \\
        3 steps   & 10.15 &   spectro & 6,651 (314) & 0,273 & 56h55' &   \textbf{28.35}  &   \textbf{28.95} \\
        2 steps   & 10.15 &   spectro & 4,417 (225) & 0.173 & 40h52' &   32.04  &   32.85 \\
        1 step   & 10.15 &   spectro & 2,407 (123) & $\mathcal{M}_2$ & 22h16' &   46.57  &   44.50 \\
        \hline
        \hline
        \cite{evain:hal-03407172} 3 steps & 12.16 & w2v2-fr & - & - & \textit{36h} & \textbf{17.25} & \textbf{16.25} \\
        3 steps & 12.16 &   w2v2-fr & 3.597 (183) & 0,550 & 36h01' &   \textbf{18.69}  &   \textbf{16.14} \\
        2 steps & 12.16 &   w2v2-fr & 2.445 (125) & 0,116 & 24h29' &   18.24  &   16.23 \\
        1 step  & 12.16 &   w2v2-fr & 2.150 (110) & $\mathcal{M}_2$ & 21h32' &   19.68  &   18.77 \\
        \hline
        \hline
        \multicolumn{8}{|c|}{\textbf{Fine-tuned features} (+100h x4 GPU $\Rightarrow$ 92.720 kWh / 4729 gCO$_2$)} \\
        \hline
        2 steps +1   & 12.16 (+318) &   w2v2-fr slu & 2.569 (131) & $\infty$ & 27h28' &   14.25  &   13.78 \\
        1 step +1   & 12.16 (+318) &   w2v2-fr slu & 2.529 (129) & $\infty$ & 27h02' &   \textbf{14.16}  &   \textbf{13.26} \\ 
        \hline
        \cite{evain:hal-03407172}$^{(*)}$ 3 steps +1 & 12.16 (+318) & w2v2-fr asr & - & - & \textit{36h} & \textbf{14.58} & \textbf{13.78} \\
        \hline
        \hline
        \multicolumn{8}{|c|}{\textbf{Transfer learning}} \\
        \hline
        1 step +PM & 12.16 &   w2v2-fr & 2.420 (123) & 0,125 & 25h04' &   18.27  &   16.61 \\
        \hline
        \hline
        \multicolumn{8}{|c|}{\textbf{Transfer learning + fine-tuned features} (+100h x4 GPU $\Rightarrow$ 92.720 kWh / 4729 gCO$_2$)} \\
        \hline
        1 step +1 +PM & 12.16 (+318) &   w2v2-fr slu & 2.026 (103) & $\mathcal{M}_2$ & 19h23' &   \textbf{13.59}  & \textbf{13.21} \\
        \hline
        \hline
        \multicolumn{8}{|c|}{\textbf{State of the art}} \\
        \hline
        \cite{pelloin:hal-03128163} & \scriptsize{(+)} & MFCC & - & - & - & \textbf{16.1} & \textbf{13.6} \\
        \cite{ghannay:hal-03372494} & 318 & w2v2-fr slu$^{(**)}$ & - & - & - & - & \textbf{11.2} \\
        \hline

    \end{tabular}
}
\end{center}
\caption{Results on the French corpus MEDIA. $^{(*)}$: fine tuning for these results was performed on ASR output. $^{(**)}$: features fine-tuned in multiple stages (ASR, SLU) on the MEDIA corpus. \scriptsize{(+)}: \small{we could not estimate as the layer size is not reported.}}
\label{tab:MEDIA-token-ICASSP2022}
\end{table*}

\subsubsection{Results on French MEDIA}

Results on the MEDIA corpus are reported in table~\ref{tab:MEDIA-token-ICASSP2022}.
In the block \textbf{Base features} we show results obtained with the same experimental conditions used for PortMEDIA.
These results confirm that a competitive SLU model can be obtained at lower training cost: the \emph{1 step} strategy, compared to the \emph{3 steps} strategy, with \emph{w2v2-fr} features, gives a 40.23\% kWh cost reduction and a 16.29\% performance loss.
In the block \textbf{Fine-tuned features} we show results obtained with features from the w2v2-fr SSL model fine-tuned on the SLU output of the MEDIA task (w2v2-fr slu).
It is interesting to note that the model trained full end-to-end (\emph{1 step +1}) obtains better results than the model trained in 2 steps.
This is due to the fact that the full end-to-end model can be trained with a more aggressive training strategy, in particular a lower dropout regularization.
These settings are not effective for the 2-steps model, intuitively because they ``erase'' the information provided by the pre-trained encoder, bringing the model far from the optimum.
Since the SSL model is fine-tuned on the same task as the final SLU model, it is not surprising that the SLU model obtains very competitive results at a lower cost: 29.69\% kWh cost reduction and 17.84\% performance gain with respect to the \emph{3 steps} strategy.
Indeed, compared to the last state-of-the-art models on MEDIA (\textbf{State of the art} in the table), our results are better than \cite{pelloin:hal-03128163}, which is the best end-to-end model so far, and not far from \cite{ghannay:hal-03372494} which used a cascade system, but our results are obtained at a lower training cost compared to the literature.
\cite{pelloin:hal-03128163} and \cite{ghannay:hal-03372494} do not mention the computational cost of their model, however from what is reported in their work, we estimate a higher computational cost than the one needed overall for our models.
\cite{ghannay:hal-03372494} in particular uses the wav2vec 318M parameter model as a component of the SLU system and performs several fine-tuning processes like the one we performed once. \\
Results reported so far prove that SSL models allow for reaching very competitive performance even with a full end-to-end training strategy (\emph{1 step}), thus for the following experiments we use only this cheaper strategy. \\
In the blocks \textbf{Transfer learning} and \textbf{Transfer learning + fine-tuned features} of table~\ref{tab:MEDIA-token-ICASSP2022}, we show respectively results obtained with transfer learning from the PortMEDIA task, and results obtained with transfer learning from the same task using features fine-tuned on the SLU MEDIA task. \\
With transfer learning alone, using a SLU model trained on PortMEDIA as starting point for training a model on MEDIA (\textbf{+PM}), we obtain a substantial improvement on the test set: 18.77 vs. 16.61, and only 2.91\% performance loss with respect to the \emph{3 steps} strategy (16.14\% CER) with 32.72\% kWh cost reduction. This could be obtained without any additional cost assuming a situation where SLU models for PortMEDIA are available in advance.
The value \emph{kWh/p} for models using \emph{w2v2-fr} features as input is computed with respect to this model (the cheapest one with this input). \\
With transfer learning and using fine-tuned features (\textbf{Transfer learning + fine-tuned features} block), while there is an improvement on the development set (14.16 vs. 13.59), there is basically no gain on the test set (13.26 vs. 13.21).
We think this is due to the fact that, since the SSL model is already fine-tuned on the SLU task, the small contribution of PortMEDIA through the transfer learning (PortMEDIA is even smaller than MEDIA) does not add much information that is not already contained in the fine-tuned features.
This model has still the advantage of being cheaper: 19h23' vs. 27h02' for training, that is 19.88\% kWh cost reduction with respect to the \emph{1 step +1} model trained with fine-tuned features but without transfer learning. The same model allows for a 43.67\% cost reduction and 18.15\% performance gain with respect to the \emph{3 steps} strategy, again assuming conditions where SLU models for PortMEDIA and fine-tuned models are available in advance.
The value \emph{kWh/p} for models using \emph{w2v2-fr slu} features is computed with respect to this model. \\
As a final remark, we would like to point out that we were not led  \emph{``down the garden path''} blindly listening stochastic parrots \cite{10.1145/3442188.3445922}, our SLU models proved to be state-of-the-art with basic input features \cite{dinarelli2020data}, and they were improved in \cite{evain:hal-03407172}, while being generic enough to tackle any sequence prediction problem from speech.


\section{Conclusions}

In this paper we analysed simple training strategies for SLU models, with the aim of reducing the computational cost of the model training phase, while keeping competitive performance.
Our results show that, through the use of pre-trained SSL models, it is possible to attain these objectives.
Adding the computational cost for fine-tuning an SSL model on the target SLU task, we obtain the second best result of the literature on MEDIA, and we equal the state-of-the-art on the FSC task, both with a full end-to-end model trained in one shot. While fine-tuning is relatively costly, overall our models are less expensive than the best state-of-the-art models.
In order to have a comprehensive understanding of how results on a given task are obtained, it would be desirable that the community adopt as a standard practice the evaluation of models also in terms of resource consumption, especially as consequence of the more and more frequent use of huge and expensive models.

\section{Acknowledgements}

This work has been realized using the HPC computational resources of GENSI - IDRIS, contract number AD011011615R1. \\
This work was partially supported by the JCJC CREMA project (\emph{Coreference REsolution into MAchine translation}), funded by the French Agency for Research (ANR), contract number ANR-21-CE23-0021-01.

\bibliographystyle{IEEEtran}
\bibliography{mybib}

\end{document}